\documentclass[11pt]{article}

\usepackage[preprint]{acl}

\usepackage{times}
\usepackage{latexsym}

\usepackage[T1]{fontenc}

\usepackage[utf8]{inputenc}
\usepackage{booktabs}
\usepackage{multirow}
\usepackage{microtype}

\usepackage{inconsolata}

\usepackage{graphicx}
\usepackage{algorithm}
\usepackage{algorithmic}
\usepackage{amsmath,amssymb}
\usepackage{xcolor}
\title{BudgetDraft: Acceptance-Aware Multi-View Training for Sparse-KV Speculative Decoding}

\author{
Liang He$^{1,\ast}$ \quad
Jingbo Wen$^{2}$ \quad
Qishi Zhan$^{3}$ \quad
Yixiong Chen$^{4}$ \\
Kangning Cui$^{5}$ \quad
Qizhen Lan$^{6}$ \quad
Xilu Wang$^{7,\ast}$ \\[4pt]
$^{1}$Shanghai Institute of Optics and Fine Mechanics \quad
$^{2}$The University of Sydney \quad
$^{3}$Marquette University \\
$^{4}$Johns Hopkins University \quad
$^{5}$Wake Forest University \\
$^{6}$University of Texas Health Science Center at Houston \quad
$^{7}$University of Surrey \\[4pt]
\texttt{hel@siom.ac.cn, wangxilu@surrey.ac.uk} \\[2pt]
$^\ast$Corresponding authors
}

\begin{document}
\maketitle

\begin{abstract}
Speculative decoding speeds up autoregressive decoding by using a drafter to propose multiple tokens that a verifier validates in parallel. In resource-constrained deployments, the drafter uses a sparse KV cache to limit peak GPU memory and end-to-end latency under a fixed KV budget, while the verifier keeps a full KV cache. Mid-to-long context inference (4K--16K context length) is common in real applications. However, naive sparse/full speculative decoding suffers from the sparse/full mismatch as context length grows, causing the acceptance rate to drop quickly.
We propose BudgetDraft, a multi-view sparse training method for sparse drafting in mid-to-long inference. The drafter is exposed to multiple sampled KV budgets during training and learns to align each sparse view with one shared full-cache teacher target. BudgetDraft combines an acceptance-aware loss on a full-cache branch with a multi-view loss on a sparse-cache branch, producing a single budget-robust drafter that recovers acceptance across sparsity levels without extra inference-time components. 
Experimental results on PG-19, LongBench, and LWM show that BudgetDraft achieves up to 6.55$\times$, 4.46$\times$, and 2.10$\times$ end-to-end speedup vs AR at 4K, 8K, and 16K context lengths, while keeping the inference pipeline memory-friendly. 
\end{abstract}

\section{Introduction}
Mid-to-long context inference, where the context length ranges from 4K to 16K tokens, is increasingly common in real-world applications such as document summarization, multi-turn dialogue, and retrieval-augmented generation~\citep{liu2025longcontext,liao2025e2llm}. Autoregressive decoding (AR) remains the dominant paradigm for large language model (LLM) inference, but its sequential token-by-token generation leads to high latency and cost~\citep{Chen2026EdgeLLM}. Speculative decoding (SD) mitigates this bottleneck by using a small drafter to propose candidate tokens, which a larger verifier validates~\citep{leviathan2023fast}. Its speedup, however, depends critically on the alignment between the drafter and the verifier. When the drafter aligns well with the verifier, multiple tokens can be accepted per verification step. 

Maintaining this alignment becomes difficult in mid-to-long inference in deployment~\citep{xiao2025duoattention}. A central constraint is the KV cache~\citep{zhou2024efficientllm}. To control peak GPU memory (VRAM) and end-to-end latency, the drafter often runs with a sparse KV cache under a fixed KV budget, retaining only a subset of cached key-value pairs, whereas the verifier keeps a full KV cache to preserve output quality~\citep{li2024kvcachesurvey}. This sparse/full mismatch grows with context length and can sharply reduce the acceptance rate. Importantly, it also makes speculative decoding highly sensitive to the KV budget, which is problematic in deployment where memory budgets vary across devices and workloads \citep{cai2026edgeinfer}.

\begin{figure}[t]
  \centering
  \includegraphics[width=0.95\columnwidth]{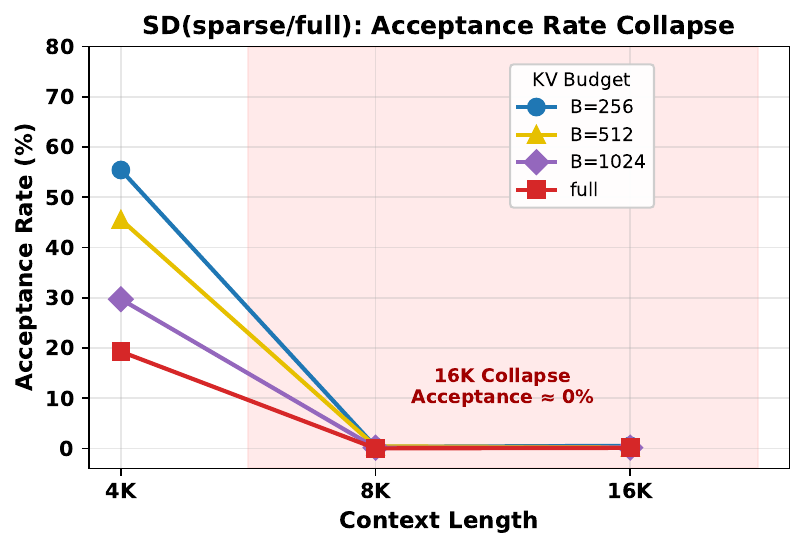}
  \caption{Acceptance collapse of SD (sparse/full) on the GS dataset with $\gamma=5$. Acceptance decreases as context length increases and becomes near-zero in the 8K--16K range across KV settings $B\in\{256,512,1024,\mathrm{full}\}$, where the full setting corresponds to 2048 tokens in this experiment.}
  \label{fig:collapse}
\end{figure}

Figure~\ref{fig:collapse} highlights this failure mode. As context length increases, SD with sparse-drafter/full-verifier settings exhibits a rapid acceptance degradation, and at 16K the acceptance becomes near-zero while the end-to-end speedup vs AR falls to $\le 1\times$. We term this the \textbf{16K collapse}. This motivates our focus on the 8K--16K range, where sparse drafting is practically attractive but acceptance degradation becomes the dominant barrier to speedup. Figure~\ref{fig:collapse} also shows that acceptance responds to the KV budget in a non-monotonic way: a smaller KV budget can yield higher acceptance than a larger one for naive sparse drafting (See more details in Appendix~\ref{app:sd_budget_paradox}).

Prior work either proposes stronger drafters under full-cache or short-context drafting~\citep{li2024eagle,li2026eagle} or sparse KV-cache and structural methods control memory or relieve long-context bottlenecks at the cost of single-model objectives or extra inference-time stages~\citep{li2024snapkv,xiao2024efficient,sun2024triforce}. Across these lines, the drafter is tuned for a fixed setting and stays brittle when the KV budget changes at deployment~\citep{gao2025aptserve}. We instead address the sparse/full mismatch at training time while keeping inference simple, and propose BudgetDraft for sparse drafting with a focus on budget robustness. Our key idea is multi-view sparse training: during training, the drafter is exposed to multiple randomly sampled KV budgets and learns to align with the same full-KV teacher target under each sparsity level. This produces a budget-robust drafter that maintains stable acceptance across KV budgets at deployment under varying memory constraints. Our contributions are:
\begin{itemize}
  \item We characterize the sparse/full acceptance rate collapse from 4K to 16K context length, establishing 16K as a practical failure boundary for naive sparse speculative decoding, and further reveal a non-monotonic budget effect at 4K where a smaller KV budget can yield higher acceptance.
  \item We propose BudgetDraft, which combines acceptance-aware alignment with multi-budget sparse-view training to produce a budget-invariant drafter that recovers acceptance rates across all sparsity levels.
  \item We demonstrate that BudgetDraft achieves up to 6.55$\times$ end-to-end speedup over AR at 4K and 4.46$\times$ at 8K on a single A100 GPU, and achieves up to $2.10\times$ speedup at 16K.
\end{itemize}

\section{Related Work}

\paragraph{Speculative Decoding.}
Speculative decoding accelerates AR inference by using a small drafter to propose tokens that a verifier checks in one forward pass~\citep{leviathan2023fast}. Recent work improves draft quality with stronger drafting mechanisms~\citep{li2024eagle}. EAGLE and its variants train lightweight draft heads from the target model's hidden states, achieving strong speedups in short contexts. EAGLE-3 adds multi-layer feature fusion and test-time simulation~\citep{li2025eagle3}. However, these methods are typically evaluated at 2K context with full-cache drafting~\citep{liu2026illusion}, and do not address the sparse/full mismatch when a sparse KV cache constrains the drafter in longer contexts~\citep{yang2025longspec}.

\paragraph{Sparse KV Cache for Mid-to-Long Context Inference.}
Many methods reduce KV cache memory by sparsification for mid-to-long inference~\citep{zhang2024h2o,li2024kvcachesurvey}. StreamingLLM retains attention sinks and recent tokens to support long-context generation with a bounded cache~\citep{xiao2024efficient}. H$_2$O uses attention-score-based eviction to keep a budgeted subset of KV states~\citep{zhang2024h2o}. A common design pattern in this line of work is budgeted eviction with token- or chunk-level selection under a fixed KV budget~\citep{zhang2024h2o,li2024snapkv,feng2025adakv}. In practical deployment, however, the KV budget is not fixed: it can vary with GPU memory availability, workload concurrency, and latency constraints~\citep{gao2025aptserve}. A drafter tuned for a single budget can therefore be brittle when deployed under different budgets. This motivates budget-robust training for sparse drafting, where the drafter can remain stable across multiple sparse views.

\paragraph{Structural Mitigation of Sparse/Full Mismatch.}
TriForce introduces a hierarchical pipeline with an intermediate retrieval-cache stage to mitigate long-context bottlenecks and improve SD under constrained drafting~\citep{sun2024triforce}. This structural approach can be effective but increases inference-time complexity by adding additional components beyond the standard drafter--verifier pipeline~\citep{sadhukhan2025magicdec}. In contrast, BudgetDraft is training-based and keeps inference simple: it trains a single drafter to remain stable across KV budgets, without introducing extra intermediate models or stages at inference~\citep{yang2025longspec}.

\begin{figure*}[t!]
  \centering
  \includegraphics[width=1\textwidth]{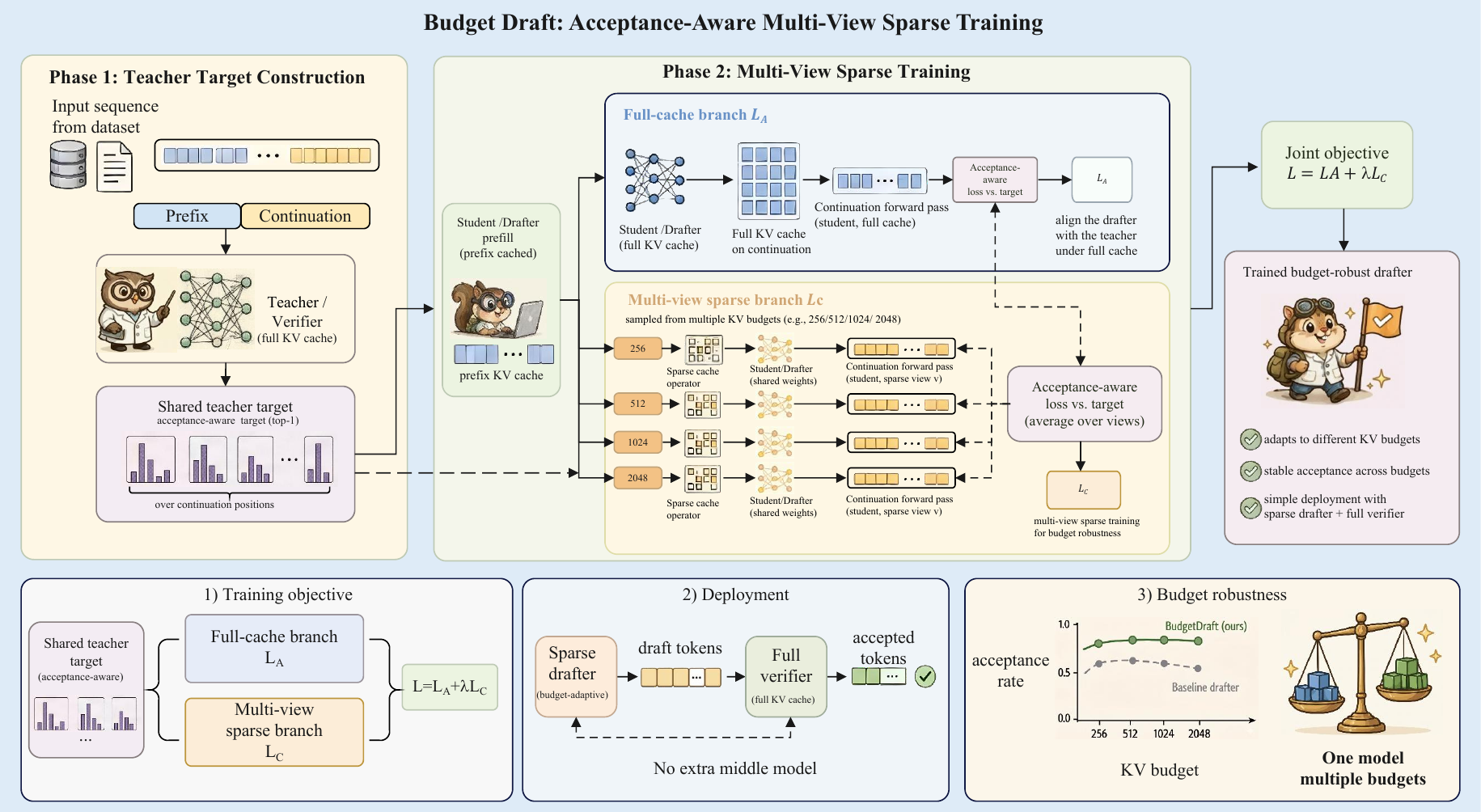}
  \caption{Overview of BudgetDraft. The verifier (teacher) produces greedy targets using a full KV cache. The drafter (student) prefills the prefix once to build a prefix KV cache, then performs two continuation forwards: a full-cache view for $\mathcal{L}_A$ and a sparse-cache view for $\mathcal{L}_C$ with a sampled KV budget $B$. Both views are supervised by the same targets, encouraging budget-robust drafting.}
  \label{fig:architecture}
\end{figure*}

\paragraph{Knowledge Distillation for Speculative Decoding.}
Beyond using off-the-shelf drafters or inference-time decoding and cache optimizations, recent work has explored improving draft models through additional training or distillation~\citep{hu2026adaspec}. Prior work studies online updates during inference, or trains task-specific drafters with distillation-style objectives~\citep{liu2024online}. Top-1 or top-$k$ distillation is a common choice for acceleration and alignment, since it focuses supervision on the verifier-preferred tokens rather than matching the full distribution~\citep{lin2025generative}. However, these approaches do not directly target the distribution shift induced by sparse KV caches, nor do they aim to keep acceptance stable across KV budgets~\citep{sadhukhan2025magicdec}. BudgetDraft uses acceptance-aware top-1 supervision together with multi-view sparse training under multi-view KV sampling, making the drafter robust to sparse-cache conditions and budget variation in deployment.

\section{Method}
\label{sec:method}
\subsection{Problem Formulation}
\label{sec:formulation}

\paragraph{Notation.}
Let $\mathcal{V}$ denote the vocabulary.
Given a token sequence $\mathbf{x}=(x_1,\dots,x_T)$, we write $\mathbf{x}_{<t}=(x_1,\dots,x_{t-1})$ for the prefix up to position $t$.
We split $\mathbf{x}$ into a prefix of length $P$ and a continuation of length $C$, so that $T=P+C$.
We supervise continuation positions $t\in\{P{+}1,\dots,T\}$.

\paragraph{Verifier (Teacher).}
The verifier is a large frozen model $M_L$ with parameters $\theta_L$.
It conditions on the full KV cache $\mathcal{K}_L^{\text{full}}$ computed from $\mathbf{x}_{<t}$ and produces the next-token distribution
\begin{equation}
  p_L(\cdot \mid \mathbf{x}_{<t};\,\mathcal{K}_L^{\text{full}}).
  \label{eq:p_L}
\end{equation}

\paragraph{Drafter (Student).}
The drafter is a small trainable model $M_S$ with parameters $\theta_S$.
At inference, it operates with a sparse KV cache $\mathcal{K}_S^{\text{sp}}$ under a KV budget $B$ :
\begin{equation}
  p_S(\cdot \mid \mathbf{x}_{<t};\,\mathcal{K}_S^{\text{sp}}(B)).
  \label{eq:p_S}
\end{equation}
Here $\mathcal{K}_S^{\text{sp}}(B)$ is obtained by selecting a budgeted subset of the drafter prefix cache.

\paragraph{Acceptance Criterion.}
Under greedy SD, a draft token at position $t$ is accepted if the drafter and verifier agree on the top-1 token:
\begin{equation}
\begin{aligned}
\hat{x}_t &= \arg\max_{v} p_S\!\left(v \mid \mathbf{x}_{<t};\,\mathcal{K}_S^{\text{sp}}(B)\right),\\
x_t^\star &= \arg\max_{v} p_L\!\left(v \mid \mathbf{x}_{<t};\,\mathcal{K}_L^{\text{full}}\right),\\
\text{accept} &\iff \hat{x}_t = x_t^\star.
\end{aligned}
\label{eq:accept}
\end{equation}
The acceptance rate $\alpha$ is the fraction of accepted draft tokens. Our goal is to increase $\alpha$ and keep it stable across KV budgets $B$ at inference.

\subsection{Overview}
\label{sec:overview}

In deployment, the drafter KV budget is not fixed, as it changes with available VRAM, workload concurrency, and latency targets. A drafter tuned for a single budget can be brittle when the budget changes, leading to unstable acceptance and unreliable speedup. This motivates training the drafter under multiple sparse views induced by different KV budgets, so that acceptance remains stable when the budget changes at inference.

As shown in Figure~\ref{fig:architecture}, BudgetDraft trains the drafter $M_S$ while keeping the verifier $M_L$ frozen, using greedy teacher targets produced under the verifier's full KV cache. The training objective combines two complementary losses:
\begin{equation}
  \mathcal{L} = \mathcal{L}_A + \lambda\,\mathcal{L}_C.
  \label{eq:total_loss}
\end{equation}
$\mathcal{L}_A$ aligns the drafter with the verifier under the full prefix cache and directly targets the greedy acceptance rule, while $\mathcal{L}_C$ performs multi-view sparse training by sampling KV budgets during training to improve robustness to inference deployment.

\subsection{Teacher Targets and Acceptance-Aware Alignment}
\label{sec:loss_a}

For each continuation position $t\in\{P{+}1,\dots,T{-}1\}$, the verifier produces a greedy teacher target
\begin{equation}
  x_t^\star = \arg\max_{v\in\mathcal{V}} p_L\!\left(v \mid \mathbf{x}_{<t};\,\mathcal{K}_L^{\text{full}}\right).
  \label{eq:teacher_target}
\end{equation}
We train the drafter to predict $x_t^\star$ under the full prefix cache using a top-1 cross-entropy loss:
\begin{equation}
  \mathcal{L}_A = -\sum_{t}\log p_S\!\left(x_t^\star \mid \mathbf{x}_{<t};\,\mathcal{K}_S^{\text{full}}\right).
  \label{eq:loss_a}
\end{equation}
It matches the greedy acceptance criterion in Eq.~\eqref{eq:accept} as it encourages the drafter to place its highest probability on the verifier's greedy token. Unlike full-distribution distillation, top-1 supervision aligns directly with the accept/reject mechanism and reduces unnecessary distribution matching.

\begin{algorithm}[t]
\small
\caption{BudgetDraft Training Procedure}
\label{alg:train}
\linespread{1.08}\selectfont
\begin{algorithmic}[1]
\setlength{\itemsep}{2pt}

\REQUIRE Verifier $M_L$ (frozen), Drafter $M_S$ (trainable)
\REQUIRE Sequence $\mathbf{x}=(x_1,\dots,x_T)$, prefix length $P$, continuation length $C=T{-}P$
\REQUIRE Budget set $B$, weights $\mathbf{w}$, chunk size $s$, loss weight $\lambda$

\vspace{0.12cm}
\STATE \textbf{Step 1: Teacher targets (no gradient)}
\STATE $
\mathcal{K}_L^{\text{full}}
\leftarrow
\text{ChunkedPrefill}(M_L,\mathbf{x}_{1:P})
$
\STATE $
\mathbf{logits}_L
\leftarrow
M_L(\mathbf{x}_{P+1:T} \mid \mathcal{K}_L^{\text{full}})
$
\STATE $
x_t^{\star}
\leftarrow
\arg\max \mathbf{logits}_L[\,t\,],
\quad
t=P{+}1,\dots,T{-}1
$
\hfill $\triangleright$ aligned to continuation positions

\vspace{0.18cm}
\STATE \textbf{Step 2: Drafter prefix prefill (no gradient)}
\STATE $
\mathcal{K}_S^{\text{full}}
\leftarrow
\text{ChunkedPrefill}(M_S,\mathbf{x}_{1:P})
$

\vspace{0.18cm}
\STATE \textbf{Step 3: Full-cache branch ($\mathcal{L}_A$)}
\STATE $
\mathbf{logits}_A
\leftarrow
M_S(
\mathbf{x}_{P+1:T}
\mid
\text{Clone}(\mathcal{K}_S^{\text{full}})
)
$
\STATE $
\mathcal{L}_A
\leftarrow
\text{CE}(\mathbf{logits}_A,\{x_t^{\star}\})
$

\vspace{0.18cm}
\STATE \textbf{Step 4: Sparse-cache branch ($\mathcal{L}_C$)}
\STATE $
B \sim \text{Cat}(B,\mathbf{w})
$
\STATE $
\mathcal{K}_S^{\text{sp}}
\leftarrow
\text{TopKChunks}(\mathcal{K}_S^{\text{full}},B,s)
$
\STATE $
\mathbf{pos}
\leftarrow
(P,P{+}1,\dots,T{-}1)
$
\hfill $\triangleright$ real position IDs
\STATE $
\mathbf{logits}_C
\leftarrow
M_S(
\mathbf{x}_{P+1:T}
\mid
\mathcal{K}_S^{\text{sp}},
\mathbf{pos}
)
$
\STATE $
\mathcal{L}_C
\leftarrow
\text{CE}(\mathbf{logits}_C,\{x_t^{\star}\})
$

\vspace{0.18cm}
\STATE \textbf{Step 5: Update}
\STATE $
\mathcal{L}
\leftarrow
\mathcal{L}_A + \lambda\,\mathcal{L}_C
$
\STATE $
\theta_S
\leftarrow
\theta_S - \eta\,\nabla_{\theta_S}\mathcal{L}
$

\end{algorithmic}
\end{algorithm}

\subsection{Multi-View Sparse Training}
\label{sec:loss_c}

Training only with $\mathcal{L}_A$ can produce a drafter that works well under the full prefix cache but degrades when deployed with a sparse KV cache. We address this gap with multi-view sparse training: at each step we sample a KV budget and train the drafter under the corresponding sparse cache, using the same greedy teacher targets $x_t^\star$ from Eq.~\eqref{eq:teacher_target}.

Concretely, we draw $B \sim \text{Cat}(B,\mathbf{w})$ with $B=\{256,512,1024,2048\}$ and $\mathbf{w}=\{0.4,0.3,0.2,0.1\}$. We construct $\mathcal{K}_S^{\text{sp}}(B)$ by partitioning the prefix cache (length $P$) into chunks of size $s=8$, scoring each chunk by cumulative attention weight, and retaining the top-$\lfloor B/s\rfloor$ chunks. The loss is then
\begin{equation}
\mathcal{L}_C =
\mathbb{E}_{B}\Bigl[
-\sum_{t}\log p_S\!\left(x_t^\star \mid \mathbf{x}_{<t};\,\mathcal{K}_S^{\text{sp}}(B)\right)
\Bigr].
\label{eq:loss_c}
\end{equation}
Sampling different budgets over training exposes the drafter to diverse sparse views of the same prefix, which reduces overfitting to a single budget and improves budget robustness at inference. After sparsification, cache length no longer matches absolute positions; we therefore pass explicit position IDs $(P,P{+}1,\dots,T{-}1)$ during the continuation forward to keep RoPE consistent with the verifier.

\begin{algorithm}[t]
\small
\caption{BudgetDraft Inference (Greedy Speculative Decoding)}
\label{alg:infer}
\begin{algorithmic}[1]
\REQUIRE Trained drafter $M_S$, verifier $M_L$
\REQUIRE Prompt $\mathbf{x}_{1:P}$, KV budget $B$, draft length $\gamma$

\STATE \textbf{Prefill:}
\STATE \quad $\mathcal{K}_L^{\text{full}} \leftarrow \text{Prefill}(M_L,\mathbf{x}_{1:P})$ \hfill $\triangleright$ full KV
\STATE \quad $\mathcal{K}_S^{\text{full}} \leftarrow \text{Prefill}(M_S,\mathbf{x}_{1:P})$
\STATE \quad $\mathcal{K}_S^{\text{sp}}(B) \leftarrow \text{TopKChunks}(\mathcal{K}_S^{\text{full}},B)$ \hfill $\triangleright$ sparsify

\REPEAT
  \STATE \textbf{Draft:} generate $\hat{\mathbf{x}}=(\hat{x}_1,\dots,\hat{x}_\gamma)$ from $M_S$ using $\mathcal{K}_S^{\text{sp}}(B)$
  \STATE \textbf{Verify:} run $M_L$ on $\hat{\mathbf{x}}$ using $\mathcal{K}_L^{\text{full}}$ and obtain verifier tokens $\mathbf{x}^\star=(x_1^\star,\dots,x_\gamma^\star)$
  \STATE Let $k$ be the largest index such that $\hat{x}_i = x_i^\star$ for all $i\le k$ \hfill $\triangleright$ Eq.~\eqref{eq:accept}
  \STATE Append $(\hat{x}_1,\dots,\hat{x}_k)$ to the output
  \STATE Append one verifier token $x_{k+1}^\star$ to the output \hfill $\triangleright$ bonus token
  \STATE Update $\mathcal{K}_L^{\text{full}}$ and $\mathcal{K}_S^{\text{sp}}(B)$ with the appended tokens
\UNTIL{max tokens reached or EOS}

\end{algorithmic}
\end{algorithm}

\subsection{Training and Inference}
\label{sec:train_infer}

\paragraph{BudgetDraft Training Procedure.}
Algorithm~\ref{alg:train} summarizes one training step. The verifier produces greedy teacher targets for the continuation positions. The drafter prefills the prefix once without gradient, then reuses the prefix cache for two continuation forwards, i.e., a full-cache branch for $\mathcal{L}_A$ and a sparse-cache branch for $\mathcal{L}_C$ with a sampled KV budget $B$. We back-propagate the combined loss through both branches and update the drafter parameters.

\paragraph{Hyperparameters.}
We train for 5,000 steps on PG-19 (streaming) with sequence length $T=16{,}384$ ($P=16{,}128$, $C=256$).
We use AdamW with learning rate $10^{-5}$, weight decay $0.01$, linear warmup for 150 steps followed by cosine decay, gradient clipping at $1.0$, and batch size 1.
We use $B=\{256,512,1024,2048\}$ with weights $\mathbf{w}=\{0.4,0.3,0.2,0.1\}$ and chunk size $s=8$ for sparse cache construction.
Training runs on a single NVIDIA A100 80GB GPU and takes approximately 5 hours.

\paragraph{BudgetDraft Inference.}
BudgetDraft follows the standard SD pipeline, summarized in Algorithm~\ref{alg:infer}. 
The trained drafter generates $\gamma$ candidate tokens autoregressively using a sparse KV cache with a chosen budget $B$.
The verifier checks all $\gamma$ candidates in a single forward pass using its full KV cache.
Accepted tokens are appended to the output; if a rejection occurs at position $k$, the verifier's token at position $k$ is used.
This inference pipeline is simple: it uses only a sparse drafter and a full verifier, and the same trained drafter can be deployed under different KV budgets as deployment constraints change.

\section{Experimental Setup}
\label{sec:experiments}

\paragraph{Models.}
We use YaRN-Llama-2-7B-128K \citep{peng2023yarn,touvron2023llama} as the verifier (6.7B parameters, fp16) and llama-68m \footnote{\url{https://huggingface.co/JackFram/llama-68m}} as the drafter (68M parameters, fp32).

\paragraph{Datasets.}
We evaluate on three datasets spanning different text domains. GS uses the PG-19 test split \citep{rae2019compressive} and contains long-form book text. LongBench uses QMSum \citep{bai2024longbench} and focuses on meeting transcript summarization. LWM uses NarrativeQA \citep{kocisky2018narrativeqa} and tests question answering over long narratives.

\paragraph{Context Lengths.}
We evaluate three context lengths: 4K (prompt length 3800), 8K (prompt length 8192), and 16K (prompt length 16384). All experiments generate 256 tokens.

\paragraph{Budgets and Draft Length.}
The verifier uses a full KV cache. The drafter uses a sparse KV cache with budget $B\in\{256,512,1024,2048\}$ and has a native maximum position embedding of 2048 tokens. We sweep the draft length $\gamma$ and report the best result over the sweep in Table~\ref{tab:best_speedup_and_accept_per_budget}, while some comparisons use a fixed $\gamma$ (e.g., $\gamma=5$).

\paragraph{Baselines.}
We compare against AR (standard autoregressive decoding), SD (sparse/full) (speculative decoding with an untrained drafter using a sparse KV cache), TriForce (hierarchical speculative decoding with a retrieval cache), and EAGLE-3 (speculative decoding with a trained draft head). Implementation details for TriForce and EAGLE-3 follow Section~\ref{sec:compare_baselines}.

\paragraph{Metrics and Hardware.}
We report acceptance rate (top-1 match under greedy decoding), end-to-end speedup vs AR (including prefill and decode time), and peak VRAM usage (see Appendix~\ref{peak_vram}). All results are averaged over 5 repeated runs under the same setting (with a warmup run discarded). We additionally analyze per-sample variance across draft lengths in Appendix~\ref{app:errorbar}. All experiments run on a single NVIDIA A100 80GB GPU.

\begin{table*}[t]
\centering
\small
\setlength{\tabcolsep}{1.4pt}
\renewcommand{\arraystretch}{1.15}

\begin{tabular}{ll ccc ccc ccc}
\toprule
\multicolumn{11}{l}{(A) Best BudgetDraft speedup (per KV budget)} \\
\midrule
& & \multicolumn{3}{c}{GS} & \multicolumn{3}{c}{LongBench} & \multicolumn{3}{c}{LWM} \\
\cmidrule(lr){3-5}\cmidrule(lr){6-8}\cmidrule(lr){9-11}
Context & Budget & AR & SD (sparse/full) & BudgetDraft & AR & SD (sparse/full) & BudgetDraft & AR & SD (sparse/full) & BudgetDraft \\
\midrule
\multirow{4}{*}{4K}
& 256  & 38.10 & 1.05$\times$/10.19 & 5.31$\times$/67.98 & 37.06 & 0.62$\times$/4.86 & 5.55$\times$/67.44 & 35.73 & 1.56$\times$/13.76 & \textcolor{red!60}{\textbf{6.54$\times$/79.37}} \\
& 512  & 38.10 & 0.84$\times$/7.36  & 5.31$\times$/67.98 & 37.06 & 0.54$\times$/4.18 & 5.56$\times$/67.44 & 35.73 & 0.92$\times$/7.49  & \textcolor{red!60}{\textbf{6.55$\times$/79.37}} \\
& 1024 & 38.10 & 0.32$\times$/1.43  & 5.32$\times$/67.98 & 37.06 & 0.36$\times$/2.06 & 5.56$\times$/67.44 & 35.73 & 0.72$\times$/5.47  & \textcolor{red!60}{\textbf{6.54$\times$/79.37}} \\
& 2048 & 38.10 & 0.30$\times$/1.17  & 5.28$\times$/67.98 & 37.06 & 0.23$\times$/0.77 & 5.56$\times$/67.44 & 35.73 & 0.24$\times$/0.86  & \textcolor{red!60}{\textbf{6.55$\times$/79.37}}\\
\midrule
\multirow{4}{*}{8K}
& 256  & 30.40 & 0.83$\times$/0.11 & 4.26$\times$/51.76 & 30.10 & 1.31$\times$/2.46 & 2.13$\times$/20.43 & 29.51 & 0.84$\times$/0.06 & \textcolor{red!60}{\textbf{4.43$\times$/55.11}} \\
& 512  & 30.40 & 0.84$\times$/0.14 & 4.27$\times$/52.26 & 30.10 & 1.27$\times$/1.73 & 2.13$\times$/20.43 & 29.51 & 0.84$\times$/0.05 & \textcolor{red!60}{\textbf{4.46$\times$/55.11}} \\
& 1024 & 30.40 & 0.83$\times$/0.05 & 4.27$\times$/52.26 & 30.10 & 1.24$\times$/1.14 & 2.13$\times$/20.43 & 29.51 & 0.84$\times$/0.05 & \textcolor{red!60}{\textbf{4.46$\times$/55.11}} \\
& 2048 & 30.40 & 0.82$\times$/0.00 & 4.27$\times$/52.26 & 30.10 & 1.23$\times$/0.97 & 2.13$\times$/20.43 & 29.51 & 0.84$\times$/0.06 & \textcolor{red!60}{\textbf{4.46$\times$/55.11}} \\
\midrule
\multirow{4}{*}{16K}
& 256  & 19.41 & 0.78$\times$/0.51 & 1.22$\times$/18.81 & 19.67 & 0.81$\times$/0.02 & 1.53$\times$/27.78 & 19.43 & 0.60$\times$/0.02 & \textcolor{red!60}{\textbf{2.10$\times$/34.17}} \\
& 512  & 19.41 & 0.77$\times$/0.19 & 1.21$\times$/18.35 & 19.67 & 0.81$\times$/0.04 & 1.53$\times$/27.98 & 19.43 & 0.68$\times$/0.03 & \textcolor{red!60}{\textbf{2.10$\times$/34.17}}  \\
& 1024 & 19.41 & 0.77$\times$/0.16 & 1.21$\times$/18.03 & 19.67 & 0.81$\times$/0.02 & 1.53$\times$/27.98 & 19.43 & 0.67$\times$/0.03 & \textcolor{red!60}{\textbf{2.10$\times$/34.17}} \\
& 2048 & 19.41 & 0.76$\times$/0.10 & 1.21$\times$/17.98 & 19.67 & 0.81$\times$/0.02 & 1.53$\times$/27.98 & 19.43 & 0.78$\times$/0.06 & \textcolor{red!60}{\textbf{2.10$\times$/34.17}}  \\
\bottomrule
\end{tabular}

\vspace{0.6em}

\begin{tabular}{ll ccc ccc ccc}
\toprule
\multicolumn{11}{l}{(B) Best BudgetDraft acceptance (per KV budget)} \\
\midrule
& & \multicolumn{3}{c}{GS} & \multicolumn{3}{c}{LongBench} & \multicolumn{3}{c}{LWM} \\
\cmidrule(lr){3-5}\cmidrule(lr){6-8}\cmidrule(lr){9-11}
Context & Budget & AR & SD (sparse/full) & BudgetDraft & AR & SD (sparse/full) & BudgetDraft & AR & SD (sparse/full) & BudgetDraft \\
\midrule
\multirow{4}{*}{4K}
& 256  & 38.10 & 2.12$\times$/55.42 & 2.95$\times$/91.62 & 37.06 & 1.83$\times$/42.39 & 3.05$\times$/92.49 & 35.73 & 2.74$\times$/73.94 & \textcolor{red!60}{\textbf{3.26$\times$/96.34}} \\
& 512  & 38.10 & 1.87$\times$/45.54 & 2.97$\times$/91.62 & 37.06 & 2.15$\times$/54.01 & 3.07$\times$/92.49 & 35.73 & 2.13$\times$/62.02 & \textcolor{red!60}{\textbf{3.26$\times$/96.34}} \\
& 1024 & 38.10 & 1.46$\times$/29.70 & 2.98$\times$/91.62 & 37.06 & 1.67$\times$/35.77 & 3.07$\times$/92.49 & 35.73 & 1.87$\times$/40.93 & \textcolor{red!60}{\textbf{3.26$\times$/96.34}} \\
& 2048 & 38.10 & 1.16$\times$/19.26 & 2.97$\times$/91.62 & 37.06 & 1.49$\times$/29.34 & 3.06$\times$/92.49 & 35.73 & 1.16$\times$/16.38 & \textcolor{red!60}{\textbf{3.27$\times$/96.34}} \\
\midrule
\multirow{4}{*}{8K}
& 256  & 30.40 & 1.18$\times$/0.30 & 3.74$\times$/74.78 & 30.10 & 1.31$\times$/2.46 & 2.13$\times$/20.43 & 29.51 & 1.19$\times$/0.17 & \textcolor{red!60}{\textbf{3.85$\times$/77.47}} \\
& 512  & 30.40 & 1.19$\times$/0.40 & 3.74$\times$/75.22 & 30.10 & 1.27$\times$/1.73 & 2.13$\times$/20.43 & 29.51 & 1.19$\times$/0.14 & \textcolor{red!60}{\textbf{3.85$\times$/77.47}} \\
& 1024 & 30.40 & 1.17$\times$/0.14 & 3.75$\times$/75.22 & 30.10 & 1.24$\times$/1.14 & 2.13$\times$/20.43 & 29.51 & 1.18$\times$/0.13 & \textcolor{red!60}{\textbf{3.85$\times$/77.47}} \\
& 2048 & 30.40 & 1.17$\times$/0.00 & 3.74$\times$/75.22 & 30.10 & 1.23$\times$/0.97 & 2.13$\times$/20.43 & 29.51 & 1.19$\times$/0.18 & \textcolor{red!60}{\textbf{3.85$\times$/77.47}} \\
\midrule
\multirow{4}{*}{16K}
& 256  & 19.41 & 0.78$\times$/0.51 & 1.22$\times$/18.81 & 19.67 & 0.81$\times$/0.02 & 1.53$\times$/27.78 & 19.43 & 0.78$\times$/0.08 & \textcolor{red!60}{\textbf{1.94$\times$/66.48}} \\
& 512  & 19.41 & 0.77$\times$/0.19 & 1.21$\times$/18.35 & 19.67 & 0.81$\times$/0.04 & 1.53$\times$/27.98 & 19.43 & 0.78$\times$/0.07 & \textcolor{red!60}{\textbf{1.89$\times$/62.15}} \\
& 1024 & 19.41 & 0.77$\times$/0.16 & 1.21$\times$/18.03 & 19.67 & 0.81$\times$/0.02 & 1.53$\times$/27.98 & 19.43 & 0.77$\times$/0.05 & \textcolor{red!60}{\textbf{1.77$\times$/55.80}} \\
& 2048 & 19.41 & 0.76$\times$/0.10 & 1.21$\times$/17.98 & 19.67 & 0.81$\times$/0.02 & 1.53$\times$/27.98 & 19.43 & 0.78$\times$/0.06 & \textcolor{red!60}{\textbf{1.52$\times$/33.92}} \\
\bottomrule
\end{tabular}

\caption{We report throughput in tok/s for AR, and speedup vs AR / acceptance rate (\%) for SD (sparse/full) and BudgetDraft. For each KV budget, Panel (A) reports the run with the best BudgetDraft speedup, and Panel (B) the run with the best BudgetDraft acceptance; both show the corresponding SD (sparse/full) run.}
\label{tab:best_speedup_and_accept_per_budget}
\end{table*}

\section{Experimental Results}
\label{sec:Experimental Results}

\subsection{Budget-Robust SD across Context Lengths}
\label{sec:main_results}

Since the drafter budget varies with VRAM and latency, a practical method should avoid relying on a single tuned budget. This section tests whether a single trained drafter remains effective when the KV budget changes at deployment, and evaluates the effectiveness of BudgetDraft across context lengths and KV budgets. 

Table~\ref{tab:best_speedup_and_accept_per_budget} summarizes results across three datasets, four KV budgets, and three context lengths. We sweep the draft length $\gamma$. Panel (A) reports, for each budget, the run with the best BudgetDraft speedup and shows the corresponding SD (sparse/full) run under the same decoding configuration. Panel (B) reports, for each budget, the run with the best BudgetDraft acceptance and shows the corresponding SD (sparse/full) run.

BudgetDraft is effective across context lengths and far less sensitive to the KV budget than SD (sparse/full). At 4K, it achieves high acceptance and strong speedup across all budgets and datasets, whereas SD (sparse/full) varies sharply with the budget. At 8K, SD (sparse/full) enters a failure regime with near-zero acceptance across budgets on GS and LWM, whereas BudgetDraft constantly recovers high acceptance on both datasets across different budgets; LongBench remains more challenging and shows lower acceptance. At 16K, SD (sparse/full) remains ineffective on GS and LongBench, while BudgetDraft still achieves non-trivial acceptance and speedup, with the strongest results on LWM. BudgetDraft reduces the need for budget-specific tuning and better supports budget-robust SD in deployment. Moreover, it achieves these gains at a peak VRAM comparable to SD (sparse/full), as reported in Appendix~\ref{peak_vram}.

\subsection{Comparison with TriForce and EAGLE-3}
\label{sec:compare_baselines}

We focus on the mid-to-long context regime (8K--16K) and compare BudgetDraft with two representative baselines, TriForce and EAGLE-3, on the LWM dataset. For a controlled comparison, all methods use the same decoding setting with draft length $\gamma=5$.
For TriForce, we follow its framework but adapt the pipeline to operate at 8K and 16K contexts, rather than using its original long-context configuration. This evaluates TriForce-style structural mitigation under the same context lengths as our main setting. For EAGLE-3, we implement the method described in its paper and train the draft head using 8K and 16K data, and then evaluate it under the same decoding protocol.

Table~\ref{tab:comparison} reports speedup results achieved by each algorithm compared with AR. At 8K, BudgetDraft reaches $2.54\times$ across all KV budgets, well above TriForce ($1.21\times$) and EAGLE-3 ($1.64\times$). At 16K, BudgetDraft remains higher across budgets, achieving $1.94\times$ at $B{=}256$, and $1.89\times$, $1.77\times$, and $1.52\times$ for $B\in\{512,1024,2048\}$, compared to $1.19\times$ for TriForce and $1.36\times$ for EAGLE-3.
Overall, BudgetDraft maintains a clear advantage at 8K--16K while supporting flexible KV-budget choices at deployment.

\begin{table}[t]
\centering
\small
\setlength{\tabcolsep}{5pt}
\renewcommand{\arraystretch}{1.15}
\begin{tabular}{lcc}
\hline
Method & Speedup & Drafter \\
\hline
\multicolumn{3}{c}{8K context (LWM, $\gamma{=}5$)} \\
\hline
AR & 1.00$\times$ & -- \\
SD (sparse/full)  & 1.19$\times$ & 68M \\
TriForce & 1.21$\times$ & 68M+7B \\
EAGLE-3 & 1.64$\times$ & draft head \\
BudgetDraft ($B{=}256$)  & \textcolor{red!60}{\textbf{2.54$\times$}} & 68M \\
BudgetDraft ($B{=}512$)  & \textcolor{red!60}{\textbf{2.54$\times$}} & 68M \\
BudgetDraft ($B{=}1024$) & \textcolor{red!60}{\textbf{2.54$\times$}} & 68M \\
BudgetDraft ($B{=}2048$) & \textcolor{red!60}{\textbf{2.54$\times$}} & 68M \\
\hline
\multicolumn{3}{c}{16K context (LWM, $\gamma{=}5$)} \\
\hline
AR & 1.00$\times$ & -- \\
SD (sparse/full)  & 0.78$\times$ & 68M \\
TriForce & 1.19$\times$ & 68M+7B \\
EAGLE-3 & 1.36$\times$ & draft head \\
BudgetDraft ($B{=}256$)  & \textcolor{red!60}{\textbf{1.94$\times$}} & 68M \\
BudgetDraft ($B{=}512$)  & \textcolor{red!60}{\textbf{1.89$\times$}} & 68M \\
BudgetDraft ($B{=}1024$) & \textcolor{red!60}{\textbf{1.77$\times$}} & 68M \\
BudgetDraft ($B{=}2048$) & \textcolor{red!60}{\textbf{1.52$\times$}} & 68M \\
\hline
\end{tabular}
\caption{Comparison with baselines on LWM at 8K and 16K with $\gamma{=}5$.}
\label{tab:comparison}
\end{table}

\subsection{Ablation: Effect of $\mathcal{L}_C$}
\label{sec:ablation}

Table~\ref{tab:ablation_lc} isolates the contribution of $\mathcal{L}_C$ by comparing $\mathcal{L}_A$ with $\mathcal{L}_A + 0.5\,\mathcal{L}_C$ under the same KV budgets, context lengths, and a fixed decoding setting ($\gamma=5$). Across datasets, adding $\mathcal{L}_C$ improves acceptance and typically increases speedup, with the most consistent gains at 4K and 8K.
At 4K, $\mathcal{L}_A + 0.5\,\mathcal{L}_C$ substantially increases acceptance and yields higher speedup across all budgets on all three datasets. At 8K, the improvement is smaller but remains consistent: speedup increases by a modest margin across budgets on GS and LWM, and LongBench also improves despite being the hardest dataset in this regime. At 16K, gains persist on GS and LongBench (acceptance and speedup both increase slightly across budgets), while LWM shows a mixed pattern where larger budgets can reduce speedup, suggesting budget-dependent trade-offs in the longest setting. To further isolate the role of multi-budget sampling, we compare it with a single-budget sparse-branch variant in Appendix~\ref{app:single_budget_ablation}. The results show that single-budget training becomes unstable under budget shifts, confirming the benefit of multi-budget sampling.

\begin{table*}[t!]
\centering
\small
\setlength{\tabcolsep}{2.5pt}
\renewcommand{\arraystretch}{1.12}
\begin{tabular}{ll ccc ccc ccc}
\toprule
& & \multicolumn{3}{c}{GS} & \multicolumn{3}{c}{LongBench} & \multicolumn{3}{c}{LWM} \\
\cmidrule(lr){3-5}\cmidrule(lr){6-8}\cmidrule(lr){9-11}
Context & Budget & AR & $\mathcal{L}_A$ & $\mathcal{L}_A + 0.5\,\mathcal{L}_C$ & AR & $\mathcal{L}_A$ & $\mathcal{L}_A + 0.5\,\mathcal{L}_C$ & AR & $\mathcal{L}_A$ & $\mathcal{L}_A + 0.5\,\mathcal{L}_C$ \\
\midrule
\multirow{4}{*}{4K}
& 256  & 38.10 & 2.63/78.43 & 2.95/91.62 & 37.06 & 3.02/90.27 & 3.05/92.49 & 35.73 & 3.23/94.50 & 3.26/96.34 \\
& 512  & 38.10 & 2.62/76.16 & 2.97/91.62 & 37.06 & 3.05/91.12 & \colorbox{blue!15}{3.07/92.49} & 35.73 & 3.25/95.12 & 3.26/96.34 \\
& 1024 & 38.10 & 2.17/57.54 & \colorbox{blue!15}{2.98/91.62} & 37.06 & 2.94/86.76 & \colorbox{blue!15}{3.07/92.49} & 35.73 & 3.20/93.04 & 3.26/96.34 \\
& 2048 & 38.10 & 1.97/49.32 & 2.97/91.62 & 37.06 & 2.53/68.88 & 3.06/92.49 & 35.73 & 2.71/72.60 & \colorbox{blue!15}{3.27/96.34} \\
\midrule
\multirow{4}{*}{8K}
& 256  & 30.40 & 2.33/71.10 & 2.39/74.78 & 30.10 & 1.34/19.30 & 1.37/20.43 & 29.51 & 2.50/75.79 & \colorbox{blue!15}{2.54/77.47} \\
& 512  & 30.40 & 2.32/70.82 & 2.39/75.22 & 30.10 & 1.31/18.27 & 1.37/20.43 & 29.51 & 2.49/75.03 & \colorbox{blue!15}{2.54/77.47} \\
& 1024 & 30.40 & \colorbox{blue!15}{2.43/77.98} & 2.40/75.22 & 30.10 & 1.26/16.44 & \colorbox{blue!15}{1.38/20.43} & 29.51 & 2.40/69.96 & \colorbox{blue!15}{2.54/77.47} \\
& 2048 & 30.40 & 2.34/72.41 & 2.39/75.22 & 30.10 & 1.23/15.40 & \colorbox{blue!15}{1.38/20.43} & 29.51 & 2.42/70.50 & \colorbox{blue!15}{2.54/77.47} \\
\midrule
\multirow{4}{*}{16K}
& 256  & 19.41 & 1.17/15.86 & \colorbox{blue!15}{1.22/18.81} & 19.67 & 1.50/27.24 & 1.53/27.78 & 19.43 & 1.91/63.79 & \colorbox{blue!15}{1.94/66.48} \\
& 512  & 19.41 & 1.16/15.73 & 1.22/18.35 & 19.67 & 1.50/27.28 & \colorbox{blue!15}{1.53/27.98} & 19.43 & 1.89/61.54 & 1.89/62.15 \\
& 1024 & 19.41 & 1.16/15.65 & 1.21/18.03 & 19.67 & 1.50/27.00 & \colorbox{blue!15}{1.53/27.98} & 19.43 & 1.88/61.60 & 1.77/55.80 \\
& 2048 & 19.41 & 1.15/14.94 & 1.21/17.98 & 19.67 & 1.50/26.63 & \colorbox{blue!15}{1.53/27.98} & 19.43 & 1.86/58.82 & 1.52/33.92 \\
\bottomrule
\end{tabular}
\caption{Ablation on multi-view sparse training at $\gamma=5$. We report throughput in tok/s for AR, and speedup vs AR / acceptance rate (\%) for $\mathcal{L}_A$ and $\mathcal{L}_A + 0.5\,\mathcal{L}_C$, under the same $\gamma$ and KV budget.}
\label{tab:ablation_lc}
\end{table*}

\subsection{Sensitivity Analysis}
\label{sec:sensitivity}

\paragraph{$\lambda$ Sensitivity.}
We compare $\mathcal{L}_A + 0.5\,\mathcal{L}_C$ ($\lambda=0.5$) with $\mathcal{L}_A + \mathcal{L}_C$ ($\lambda=1.0$). The two settings achieve very similar acceptance and speedup at 4K and 8K across datasets and KV budgets. At 16K, differences remain small and are sometimes budget-dependent. We use $\lambda=0.5$ as the default. Detailed results and analysis are provided in Appendix~\ref{app:lambda_sensitivity}.

\begin{figure*}[t!]
  \centering
  \includegraphics[width=0.98\textwidth]{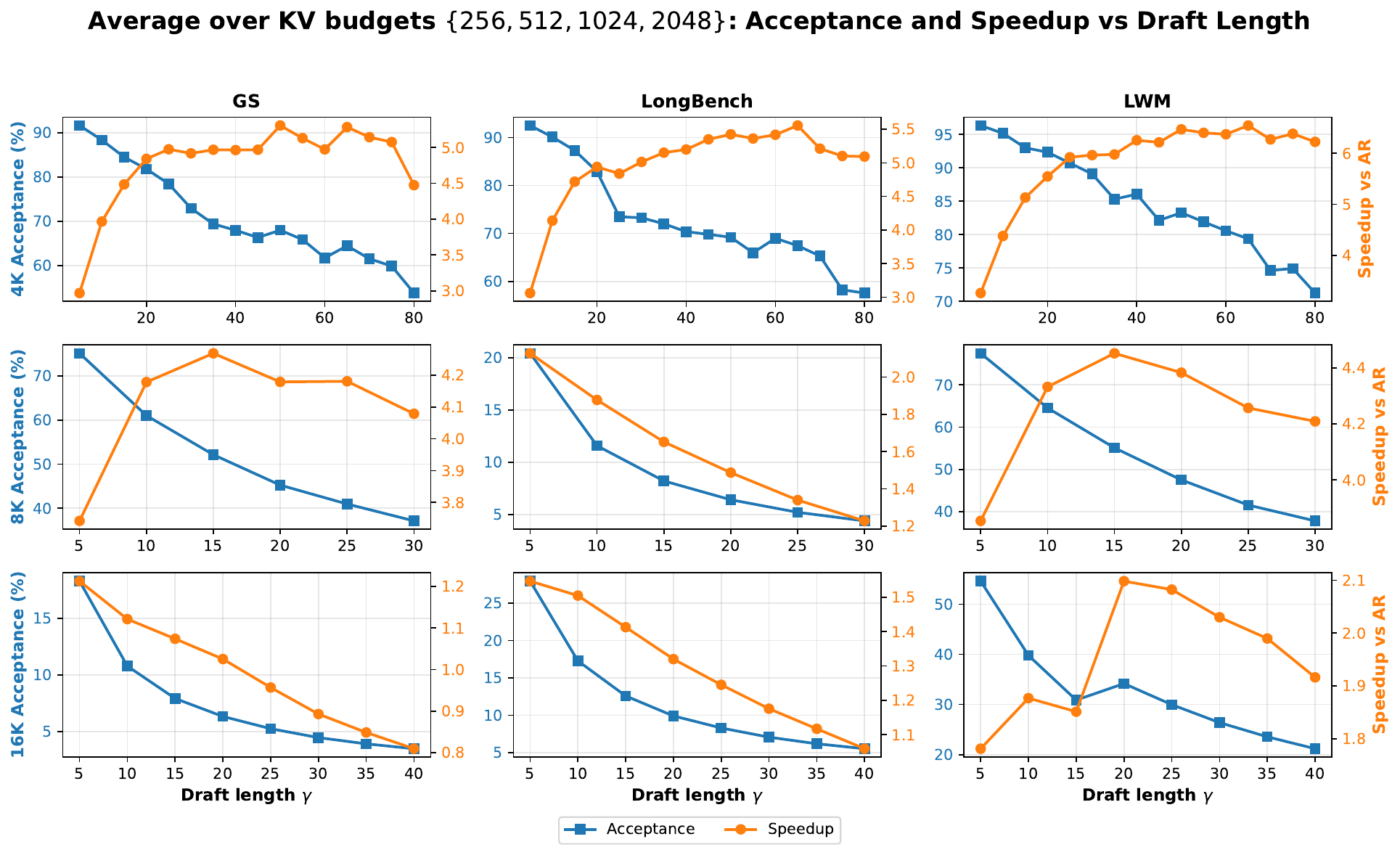}
  \caption{Budget-averaged sensitivity to draft length. We average over KV budgets $B\in\{256,512,1024,2048\}$ and plot acceptance rate and speedup vs AR as a function of draft length $\gamma$. Rows correspond to context lengths (4K/8K/16K), and columns correspond to datasets (GS/LongBench/LWM).}
  \label{fig:gamma_sensitivity_budget_avg}
\end{figure*}

\paragraph{Draft Length Sensitivity.}
Figure~\ref{fig:gamma_sensitivity_budget_avg} summarizes how acceptance and speedup change with the draft length $\gamma$ after averaging over KV budgets. At 4K, acceptance remains high across datasets for a wide range of $\gamma$, while speedup generally increases with $\gamma$ before saturating or mildly fluctuating. At 8K, GS and LWM still maintain strong acceptance over a broad range of $\gamma$, leading to clear speedup peaks at moderate $\gamma$, while LongBench shows lower acceptance and a weaker speedup peak. At 16K, acceptance drops and becomes more sensitive to $\gamma$, narrowing the range where speculative decoding is efficient; the best speedup shifts to shorter or moderate drafts depending on the dataset. These trends suggest that $\gamma$ should be tuned with context length, while the stable curves under budget averaging reflect the robustness goal of BudgetDraft in deployment.

\section{Conclusion}
\label{sec:conclusion}

We identified a sparse/full acceptance collapse in mid-to-long context speculative decoding: as the context length increases from 4K to 16K, naive sparse drafting can degrade to near-zero acceptance, making SD ineffective. We proposed BudgetDraft, a training-based alignment method that combines acceptance-aware alignment ($\mathcal{L}_A$) with multi-view sparse training ($\mathcal{L}_C$). By training the drafter under multiple randomly sampled KV budgets, BudgetDraft produces a budget-robust drafter that maintains stable acceptance and speedup under different sparsity levels at inference.

Experiments on three datasets show that BudgetDraft delivers strong end-to-end speedups in the 4K–16K regime on a single A100 GPU, while using only a lightweight 68M drafter. BudgetDraft keeps the inference pipeline simple and memory-friendly, making it practical for deployment across devices with varying memory budgets. Future work includes improving robustness at longer contexts and exploring tighter integration with sparse-cache policies.

\section*{Limitations}
\label{sec:limitations}

\paragraph{Drafter position range.}
A key limitation is the drafter's limited native position range. The 68M drafter operates with a short maximum position embedding, while our target setting spans mid-to-long contexts (8K–16K). We pass explicit position IDs to keep the drafter aligned with the verifier, but the drafter still enters a positional extrapolation regime at longer contexts, which can reduce acceptance and thus limit speedup. However, positional extrapolation alone does not fully explain the observed behavior, since BudgetDraft substantially improves acceptance under the same drafter architecture and position range. Using a drafter with a longer native context window or applying position-extension techniques is a promising direction for future work.

\paragraph{Verifier-specific training.}
BudgetDraft trains a drafter against a fixed verifier, and retraining is required when the verifier changes. While the training cost is moderate in our setup, we do not systematically study transfer across verifiers, such as reusing a drafter with lightweight adaptation. Understanding how well a trained drafter transfers to related verifier checkpoints is an important next step.

\paragraph{Greedy decoding setting.}
Our experiments focus on greedy speculative decoding to enable output-identical generation and stable speed measurements. Extending BudgetDraft to sampling-based decoding (e.g., temperature sampling) would require adapting both the training objective and the acceptance rule to handle stochasticity, which we leave for future work.

\paragraph{Scope of sparse cache policies.}
We adopt a specific budgeted sparse KV construction based on chunk-level selection. BudgetDraft is designed to be compatible with different sparse cache policies, but we do not exhaustively evaluate alternative policies or system-level implementations (e.g., kernel-level efficiency under different eviction rules). A broader study of sparse cache choices and their interaction with training objectives is left for future work.

\section*{Acknowledgments}

We thank the anonymous reviewers for their constructive feedback.

\bibliography{custom}

\appendix

\section{Peak VRAM Usage}
\label{peak_vram}
We report peak GPU memory (VRAM) using the logged field peak\_gpu\_mb and convert it to GB.
Table~\ref{tab:peak_vram_gamma_full} summarizes peak VRAM as a function of draft length $\gamma$ at 4K/8K/16K.
For each context length and $\gamma$, we average peak VRAM over KV budgets $B\in\{256,512,1024,2048\}$ and over the three datasets.
This table shows the memory trend as $\gamma$ increases under each context length, and directly compares SD (sparse/full) (original) with BudgetDraft ($\mathcal{L}_A + 0.5\,\mathcal{L}_C$).

Peak VRAM is comparable between BudgetDraft and SD (sparse/full).
At 8K and 16K, peak VRAM is nearly identical across methods for the evaluated $\gamma$ range, suggesting that peak usage is dominated by the full-KV verifier side in this sparse/full pipeline.
At 4K, SD (sparse/full) exhibits a clear increase in peak VRAM as $\gamma$ grows, while BudgetDraft remains almost flat, indicating that BudgetDraft achieves higher speedup without additional memory overhead at larger draft lengths.

For a fair, fixed decoding setting, Table~\ref{tab:peak_vram} further reports detailed peak VRAM at $\gamma=5$ for each KV budget and dataset.
The detailed results are consistent with the aggregate trend: BudgetDraft matches SD (sparse/full) at 8K/16K and is slightly lower at 4K under some budgets.

\begin{table*}[t]
\centering
\small
\setlength{\tabcolsep}{5.0pt}
\renewcommand{\arraystretch}{1.12}
\begin{tabular}{c cc cc cc}
\toprule
$\gamma$ & \multicolumn{2}{c}{4K} & \multicolumn{2}{c}{8K} & \multicolumn{2}{c}{16K} \\
\cmidrule(lr){2-3}\cmidrule(lr){4-5}\cmidrule(lr){6-7}
& SD (sparse/full) & BudgetDraft & SD (sparse/full) & BudgetDraft & SD (sparse/full) & BudgetDraft \\
\midrule
5  & 17.29 & 17.03 & 17.25 & 17.25 & 21.44 & 21.44 \\
10 & 17.73 & 17.03 & 17.25 & 17.25 & 21.44 & 21.44 \\
15 & 18.16 & 17.04 & 17.25 & 17.25 & 21.44 & 21.44 \\
20 & 18.76 & 17.04 & 17.26 & 17.26 & --    & --    \\
25 & 19.54 & 17.11 & 17.26 & 17.26 & --    & --    \\
30 & 20.04 & 17.08 & 17.26 & 17.26 & --    & --    \\
35 & 21.36 & 17.08 & --    & --    & --    & --    \\
40 & 22.25 & 17.06 & --    & --    & --    & --    \\
45 & 23.40 & 17.07 & --    & --    & --    & --    \\
50 & 24.50 & 17.06 & --    & --    & --    & --    \\
55 & 25.28 & 17.08 & --    & --    & --    & --    \\
60 & 26.99 & 17.09 & --    & --    & --    & --    \\
65 & 28.08 & 17.09 & --    & --    & --    & --    \\
70 & 28.70 & 17.09 & --    & --    & --    & --    \\
75 & 29.92 & 17.11 & --    & --    & --    & --    \\
80 & 30.89 & 17.19 & --    & --    & --    & --    \\
\bottomrule
\end{tabular}
\caption{Peak VRAM (GB) vs draft length $\gamma$. For each context length and $\gamma$, values are averaged over KV budgets $B\in\{256,512,1024,2048\}$ and the three datasets (GS, LongBench, and LWM). Missing entries indicate that $\gamma$ was not evaluated at that context length. Values are converted from \texttt{peak\_gpu\_mb}.}
\label{tab:peak_vram_gamma_full}
\end{table*}

\begin{table*}[t]
\centering
\small
\setlength{\tabcolsep}{4.5pt}
\renewcommand{\arraystretch}{1.15}
\begin{tabular}{ll|cc|cc|cc}
\toprule
& & \multicolumn{2}{c|}{GS} & \multicolumn{2}{c|}{LongBench} & \multicolumn{2}{c}{LWM} \\
\cmidrule(lr){3-4}\cmidrule(lr){5-6}\cmidrule(lr){7-8}
Context & $B$ & SD (sparse/full) & BudgetDraft & SD (sparse/full) & BudgetDraft & SD (sparse/full) & BudgetDraft \\
\midrule
\multirow{4}{*}{4k}
 & 256  & 17.17 & 17.03 & 17.19 & 17.03 & 16.99 & 16.99 \\
 & 512  & 17.36 & 17.04 & 17.11 & 17.04 & 17.03 & 16.99 \\
 & 1024 & 17.40 & 17.05 & 17.29 & 17.05 & 17.34 & 17.00 \\
 & 2048 & 17.56 & 17.07 & 17.35 & 17.07 & 17.68 & 17.03 \\
\midrule
\multirow{4}{*}{8k}
 & 256  & 17.25 & 17.25 & 17.25 & 17.25 & 17.24 & 17.24 \\
 & 512  & 17.25 & 17.25 & 17.25 & 17.25 & 17.24 & 17.24 \\
 & 1024 & 17.25 & 17.25 & 17.25 & 17.25 & 17.24 & 17.24 \\
 & 2048 & 17.25 & 17.25 & 17.25 & 17.25 & 17.24 & 17.24 \\
\midrule
\multirow{4}{*}{16k}
 & 256  & 21.44 & 21.44 & 21.44 & 21.44 & 21.42 & 21.42 \\
 & 512  & 21.44 & 21.44 & 21.44 & 21.44 & 21.42 & 21.42 \\
 & 1024 & 21.44 & 21.44 & 21.44 & 21.44 & 21.42 & 21.42 \\
 & 2048 & 21.44 & 21.44 & 21.44 & 21.44 & 21.42 & 21.42 \\
\bottomrule
\end{tabular}
\caption{Peak VRAM usage (GB) at $\gamma=5$. We report peak GPU memory converted from \texttt{peak\_gpu\_mb} under the same context length and KV budget. SD (sparse/full) uses the untrained drafter; BudgetDraft uses $\mathcal{L}_A + 0.5\,\mathcal{L}_C$.}
\label{tab:peak_vram}
\end{table*}

\section{Per-Sample Variance Across Draft Lengths}
\label{app:errorbar}

\begin{figure*}[t]
  \centering
  \includegraphics[width=0.98\textwidth]{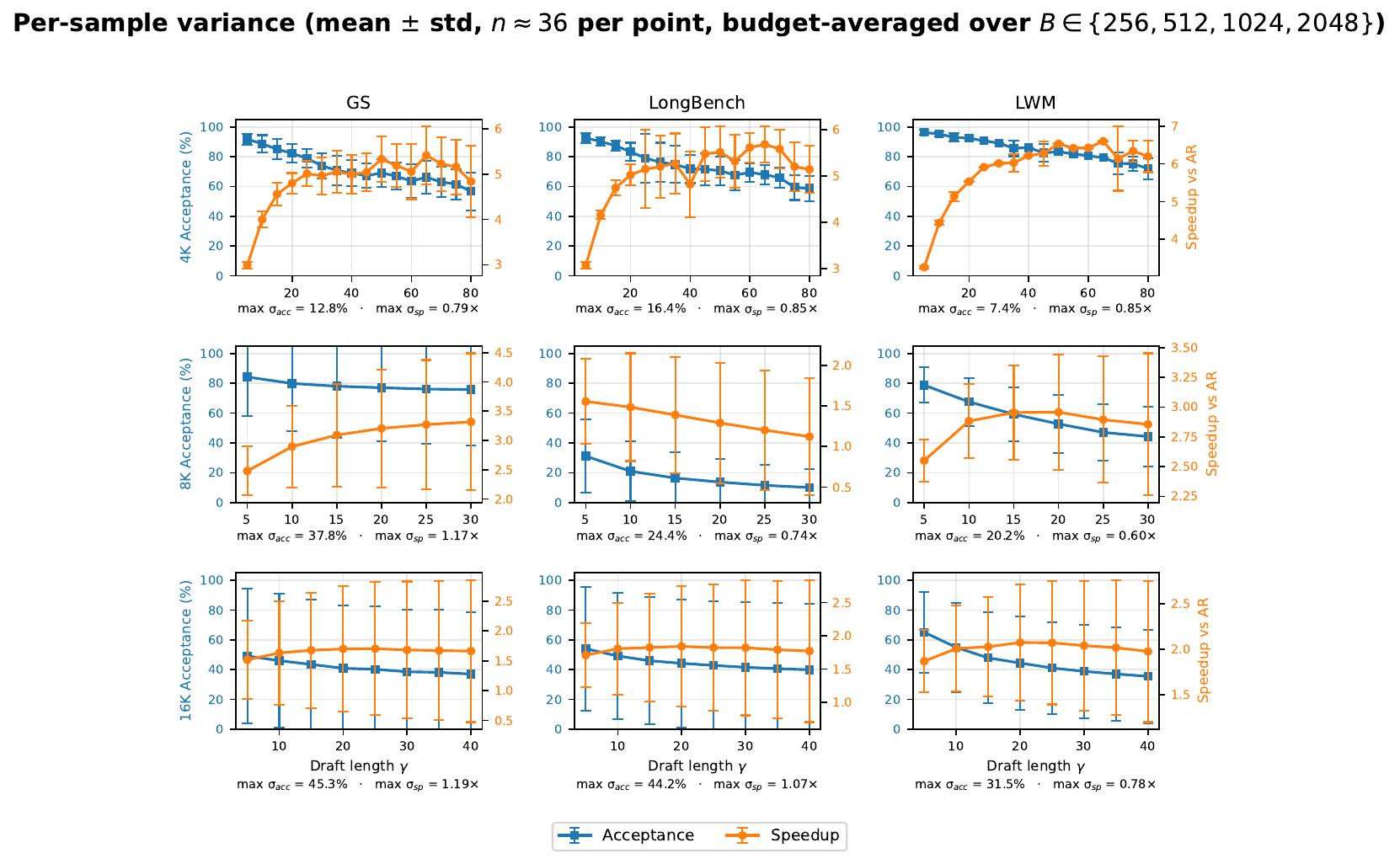}
  \caption{Per-sample variance (mean $\pm$ std) across draft lengths $\gamma$. Acceptance and speedup are averaged over KV budgets $B\in\{256,512,1024,2048\}$. Rows correspond to context lengths (4K/8K/16K), and columns correspond to datasets (GS/LongBench/LWM).}
  \label{fig:errorbar}
\end{figure*}

To evaluate the stability of speculative decoding under different draft lengths, we measure per-sample variance across datasets and context lengths. Figure~\ref{fig:errorbar} reports the mean and standard deviation of acceptance rate and end-to-end speedup versus AR as functions of draft length $\gamma$. Results are averaged over KV budgets $B\in\{256,512,1024,2048\}$, with approximately 36 samples per point.

Several trends are consistent across datasets and context lengths. First, acceptance generally decreases as $\gamma$ increases, while speedup initially improves and then saturates or declines, reflecting the standard speculative decoding trade-off between drafting more tokens and verification overhead. Second, variance remains relatively small in the main operating region, indicating that the observed gains are stable across samples rather than driven by a few outliers.

The variance patterns also differ across datasets. GS and LongBench show smoother acceptance degradation as $\gamma$ increases, whereas LWM exhibits stronger sensitivity at longer contexts, especially at 16K. This behavior is consistent with the larger semantic diversity and longer dependency structure in narrative-style inputs. Despite this increased variance, the overall trends remain stable across budgets and datasets.

Importantly, the curves remain smooth even at 8K--16K, without abrupt collapse across neighboring $\gamma$ values. This suggests that the degradation in long-context speculative decoding is not solely caused by catastrophic positional extrapolation failure. Instead, the behavior is more consistent with gradually increasing sparse/full mismatch under longer contexts, which aligns with the motivation behind BudgetDraft.

\section{$\lambda$ Sensitivity}
\label{app:lambda_sensitivity}

This appendix provides detailed results for the $\lambda$ sensitivity study in the main text. We compare $\mathcal{L}_A + 0.5\,\mathcal{L}_C$ ($\lambda=0.5$) with $\mathcal{L}_A + \mathcal{L}_C$ ($\lambda=1.0$). The complete results are reported in Table~\ref{tab:lambda_sensitivity_full}.

Across 4K and 8K, the two settings are closely matched across datasets and KV budgets. Acceptance is nearly identical, and speedup differs only marginally, indicating that performance in the main mid-context regime does not hinge on a narrow choice of $\lambda$ once the drafter is trained with multi-view sparse supervision.

At 16K, $\lambda=1.0$ is often slightly better, but the gains remain small. On GS and LongBench, $\lambda=1.0$ consistently yields a modest increase in speedup across budgets, with acceptance remaining nearly unchanged. On LWM, $\lambda=1.0$ improves speedup for most budgets, while a small exception appears at $B{=}512$, where $\lambda=0.5$ is marginally higher. This indicates that increasing the sparse-view weight can shift the trade-off across budgets in the longest setting, but the overall sensitivity is limited.

We use $\lambda=0.5$ as the default since it matches $\lambda=1.0$ closely in the main 4K--8K regime and provides a consistent choice across budgets, while avoiding the need to re-tune $\lambda$ for different deployment constraints.

\begin{table*}[t]
\centering
\small
\setlength{\tabcolsep}{2.5pt}
\renewcommand{\arraystretch}{1.12}
\begin{tabular}{ll ccc ccc ccc}
\toprule
& & \multicolumn{3}{c}{GS} & \multicolumn{3}{c}{LongBench} & \multicolumn{3}{c}{LWM} \\
\cmidrule(lr){3-5}\cmidrule(lr){6-8}\cmidrule(lr){9-11}
Ctx & Bud & AR & $\mathcal{L}_A$+0.5$\mathcal{L}_C$ & $\mathcal{L}_A$+$\mathcal{L}_C$ & AR & $\mathcal{L}_A$+0.5$\mathcal{L}_C$ & $\mathcal{L}_A$+$\mathcal{L}_C$ & AR & $\mathcal{L}_A$+0.5$\mathcal{L}_C$ & $\mathcal{L}_A$+$\mathcal{L}_C$ \\
\midrule
\multirow{4}{*}{4K}
&256  &38.10 &2.95$\times$/91.62 &2.98$\times$/91.62 &37.06 &3.05$\times$/92.49 &3.04$\times$/92.49 &35.73 &3.26$\times$/96.34 &3.25$\times$/96.34 \\
&512  &38.10 &2.97$\times$/91.62 &2.98$\times$/91.62 &37.06 &3.07$\times$/92.49 &3.06$\times$/92.49 &35.73 &3.26$\times$/96.34 &3.27$\times$/96.34 \\
&1024 &38.10 &2.98$\times$/91.62 &2.98$\times$/91.62 &37.06 &3.07$\times$/92.49 &3.06$\times$/92.49 &35.73 &3.26$\times$/96.34 &3.27$\times$/96.34 \\
&2048 &38.10 &2.97$\times$/91.62 &2.98$\times$/91.62 &37.06 &3.06$\times$/92.49 &3.06$\times$/92.49 &35.73 &3.27$\times$/96.34 &3.27$\times$/96.34 \\
\midrule
\multirow{4}{*}{8K}
&256  &30.40 &2.39$\times$/74.78 &2.23$\times$/74.56 &30.10 &1.37$\times$/20.43 &1.37$\times$/20.43 &29.51 &2.54$\times$/77.47 &2.54$\times$/77.47 \\
&512  &30.40 &2.39$\times$/75.22 &2.39$\times$/75.22 &30.10 &1.37$\times$/20.43 &1.37$\times$/20.43 &29.51 &2.54$\times$/77.47 &2.54$\times$/77.47 \\
&1024 &30.40 &2.40$\times$/75.22 &2.40$\times$/75.22 &30.10 &1.38$\times$/20.43 &1.37$\times$/20.43 &29.51 &2.54$\times$/77.47 &2.54$\times$/77.47 \\
&2048 &30.40 &2.39$\times$/75.22 &2.40$\times$/75.22 &30.10 &1.38$\times$/20.43 &1.37$\times$/20.43 &29.51 &2.54$\times$/77.47 &2.54$\times$/77.47 \\
\midrule
\multirow{4}{*}{16K}
&256  &19.41 &1.22$\times$/18.81 &1.25$\times$/18.72 &19.67 &1.53$\times$/27.78 &1.55$\times$/27.83 &19.43 &1.94$\times$/66.48 &1.99$\times$/66.97 \\
&512  &19.41 &1.22$\times$/18.35 &1.24$\times$/18.46 &19.67 &1.53$\times$/27.98 &1.55$\times$/27.98 &19.43 &1.89$\times$/62.15 &1.88$\times$/61.59 \\
&1024 &19.41 &1.21$\times$/18.03 &1.24$\times$/18.21 &19.67 &1.53$\times$/27.98 &1.55$\times$/27.98 &19.43 &1.77$\times$/55.80 &1.86$\times$/55.67 \\
&2048 &19.41 &1.21$\times$/17.98 &1.24$\times$/18.16 &19.67 &1.53$\times$/27.98 &1.55$\times$/27.98 &19.43 &1.52$\times$/33.92 &1.56$\times$/34.81 \\
\bottomrule
\end{tabular}
\caption{$\lambda$ sensitivity at $\gamma=5$ ($\mathcal{L}_A + 0.5\,\mathcal{L}_C$ vs $\mathcal{L}_A + \mathcal{L}_C$). Entries report speedup vs AR / acceptance rate (\%) under the same $\gamma$, context length, and KV budget.}
\label{tab:lambda_sensitivity_full}
\end{table*}

\section{Why Smaller KV Budgets Can Increase Acceptance at 4K}
\label{app:sd_budget_paradox}

Figure~\ref{fig:collapse} and Table~\ref{tab:best_speedup_and_accept_per_budget} show a seemingly counter-intuitive pattern for SD (sparse/full) at 4K: smaller KV budgets can lead to higher acceptance than larger budgets. This effect does not contradict the sparse/full mismatch story. Instead, it reflects how budgeted sparse KV selection interacts with a small drafter in the short-to-mid context regime.

\paragraph{Sparse KV as an information filter.}
At 4K, next-token predictions are often dominated by a small subset of the prefix, typically recent tokens and a few highly relevant segments. A smaller KV budget forces the sparse cache to keep only the most relevant chunks (or tokens) under the selection policy. This can act as an information filter that removes weakly relevant history. For a small drafter, reducing low-signal context can stabilize attention, sharpen the conditional distribution, and increase the probability that the drafter's top-1 token matches the verifier's top-1 token, which directly increases greedy acceptance.

\paragraph{More cached context can introduce distractors for a small drafter.}
While a larger KV budget intuitively adds more context, in practice it also retains much irrelevant or noisy content alongside the useful evidence. A small drafter has limited capacity to integrate long context and can be more sensitive to distractors than the verifier. With a larger sparse cache, attention mass is spread across more tokens, and small differences between drafter and verifier in how evidence is weighted can shift probability mass among competing candidates. Since greedy acceptance requires a strict top-1 match, these shifts can reduce acceptance even when overall modeling quality does not degrade.

\paragraph{Position handling near the drafter limit.}
In our setup, the drafter has a native maximum position embedding of 2048 tokens. At 4K prompts, position handling (e.g., explicit position IDs, clamping, or re-indexing for sparse-cache evaluation) can introduce additional mismatch, and this mismatch can be amplified when more distant tokens are retained. Smaller budgets tend to retain more recent or higher-attention chunks, reducing exposure to position-related distortion and making drafter--verifier agreement easier. We discuss the broader impact of position range in Section~\ref{sec:limitations}.

\paragraph{Takeaway.}
The 4K behavior is therefore expected: a smaller sparse cache can improve SD (sparse/full) acceptance by filtering noise and reducing sensitivity to long-range distractors and position-related mismatch. As context length grows, sparse/full mismatch becomes the dominant factor, and SD (sparse/full) acceptance collapses across budgets, which motivates BudgetDraft.

\section{Single-Budget vs. Multi-Budget Sparse Training}
\label{app:single_budget_ablation}

To isolate the effect of multi-budget sparse training, we compare BudgetDraft with a single-budget variant. The single-budget variant uses the same objective $\mathcal{L}_A + 0.5\,\mathcal{L}_C$, but fixes the sparse-cache branch to $B{=}1024$ during training. We evaluate both models on LWM at 16K with $\gamma=20$, where SD(sparse/full) collapses and BudgetDraft achieves its strongest 16K speedup.

\begin{table*}[t!]
\centering
\small
\setlength{\tabcolsep}{5pt}
\renewcommand{\arraystretch}{1.5}
\begin{tabular}{lcccc}
\toprule
Evaluating setting & $B{=}256$ & $B{=}512$ & $B{=}1024$ & $B{=}2048$ \\
\midrule
Single-budget ($B{=}1024$) 
& 2.10$\times$/34.17 
& 2.02$\times$/31.03 
& 1.84$\times$/24.48 
& 1.47$\times$/14.61 \\
Multi-budget 
& \colorbox{blue!15}{2.10$\times$/34.17} 
& \colorbox{blue!15}{2.10$\times$/34.17}  
& \colorbox{blue!15}{2.10$\times$/34.17} 
& \colorbox{blue!15}{2.10$\times$/34.17}  \\
\bottomrule
\end{tabular}
\caption{Single-budget vs. multi-budget sparse training on LWM at 16K with $\gamma=20$. Entries report speedup vs AR / acceptance rate (\%). The single-budget variant is trained with a fixed sparse-cache budget $B{=}1024$, while multi-budget training samples from $B\in\{256,512,1024,2048\}$.}
\label{tab:single_budget_ablation}
\end{table*}

Table~\ref{tab:single_budget_ablation} shows that single-budget sparse training does not provide stable generalization across inference budgets. Although the single-budget variant performs well at $B{=}256$, its speedup and acceptance degrade as the inference budget increases. The degradation is most pronounced at $B{=}2048$, where speedup drops to $1.47\times$ and acceptance falls to 14.61\%. In contrast, multi-budget training maintains $2.10\times$ speedup and 34.17\% acceptance across all budgets. This confirms that the benefit of $\mathcal{L}_C$ is not only from adding a sparse-cache branch, but also from exposing the drafter to multiple sparse views during training, which improves budget stability at deployment.

\end{document}